\documentclass[twocolumn,letterpaper]{IEEEtran}  % only supports two-column, letterpaper format
\usepackage[]{graphicx}    % We use this package in this document
\usepackage{amsmath}
\usepackage{booktabs}
\usepackage[font=bf, figurename=Figure, labelsep=period]{caption}
\usepackage[]{svg}
\usepackage[ruled,vlined]{algorithm2e}
\usepackage{subcaption}
\usepackage{float}
\usepackage{parskip}% http://ctan.org/pkg/parskip
\usepackage{bbm}

\usepackage[noadjust]{cite}
\newcommand{\ignore}[1]{}  % {} empty inside = %% comment

\begin{document}
\title{Mars Terrain Segmentation with Less Labels}

\author{%
Edwin Goh$^{1}$, Jingdao Chen$^{2}$, Brian Wilson$^{1}$\\
$^{1}$Jet Propulsion Laboratory, California Institute of Technology\\
$^{2}$Computer Science and Engineering, Mississippi State University\\
edwin.y.goh@jpl.nasa.gov, chenjingdao@cse.msstate.edu, bdwilson@jpl.nasa.gov
\thanks{\footnotesize \copyright2022. All rights reserved.}
}

\maketitle

\thispagestyle{plain}
\pagestyle{plain}

\maketitle
\thispagestyle{plain}
\pagestyle{plain}

\begin{abstract}

Planetary rover systems need to perform terrain segmentation to identify drivable areas as well as identify specific types of soil for sample collection. The latest Martian terrain segmentation methods rely on supervised learning which is very data hungry and difficult to train where only a small number of labeled samples are available. Moreover, the semantic classes are defined differently for different applications (e.g., rover traversal vs. geological) and as a result the network has to be trained from scratch each time, which is an inefficient use of resources. This research proposes a semi-supervised learning framework for Mars terrain segmentation where a deep segmentation network trained in an unsupervised manner on unlabeled images is transferred to the task of terrain segmentation trained on few labeled images. The network incorporates a backbone module which is trained using a contrastive loss function and an output atrous convolution module which is trained using a pixel-wise cross-entropy loss function. Evaluation results using the metric of segmentation accuracy show that the proposed method with contrastive pretraining outperforms plain supervised learning by 2\%-10\%. Moreover, the proposed model is able to achieve a segmentation accuracy of 91.1\% using only 161 training images (1\% of the original dataset) compared to 81.9\% with plain supervised learning.

\end{abstract}

\tableofcontents

\section{Introduction}
\label{sec:intro}
Planetary rovers represent one of the primary means by which humans are able to explore and understand other celestial bodies, as exemplified by the Mars 2020 rover mission \cite{williford2018} and the Lunar 2023 VIPER mission \cite{colaprete2019}. For these extraterrestrial rover missions aimed at planetary exploration and navigation, automated scene understanding in the form of terrain segmentation is important to improve the automation and reactivity for the planetary rovers, ultimately increasing the travel distance for rovers to explore unknown planetary environments. The task of terrain segmentation is defined as the process of automatically distinguishing between different types of terrain features on the planet's surface such as sand, bedrock, and large rocks from sensor data. In particular, terrain segmentation can be considered a prerequisite step for carrying out obstacle avoidance, traversability estimation, recognizing different types of rocks for sample collection, and short-term path planning of autonomous robotic systems \cite{lu2009}. Terrain segmentation is also relevant for energy-aware planning tasks such as slip prediction and minimization of travel energy \cite{maars2020}.

The primary challenge with training a model to perform terrain segmentation from planetary image data is that data acquisition is expensive and time-consuming, and there is a general lack of annotated training data, particularly for annotation tasks that require specific expert knowledge. In addition, planetary image datasets have issues with taxonomy inconsistencies, inconsistencies in segmentation detail, and inconsistencies in scale and lighting. Existing work in image-based terrain segmentation usually rely on deep supervised learning, which requires a large amount of training data to achieve a satisfactory level of accuracy \cite{spoc2016, tunet2019}. Purely supervised learning also has other downsides such as not transferring well between datasets \cite{ltf2020}. More recent research efforts have explored the use of transfer learning or representation learning methods to transfer or extract useful information from image data even when a large number of labels is not available. However, these methods have limitations when applied to terrain segmentation; for instance, SimCLR \cite{simclr2020,simclrv2} and CPC \cite{cpc2019} are developed and benchmarked on the classification task instead of the segmentation task.

In this research, we propose to leverage transfer learning and semi-supervised deep learning techniques to improve upon current state-of-the-art performance on terrain segmentation using publicly available mission data in the form of rover images from the surface of Mars. The proposed framework involves a deep segmentation network trained in an unsupervised manner on a large set of unlabeled images, and then transferred to the task of terrain segmentation trained on a small set of labeled images. In particular, this framework allows the model to make full use of the large number of unlabeled images, common for planetary data, through contrastive pretraining. Contrastive pretraining has been shown to enable models 1) to be trained using fewer labels, 2) to generalize well to different image-based applications, and 3) to outperform standard supervised learning given comparable network architectures \cite{simclr2020, dino2021}. In the proposed work, we use a network architecture that incorporates a backbone module which is trained using a contrastive loss function and an output atrous convolution module which is trained using a pixel-wise cross-entropy loss function. The proposed method is compared against a baseline supervised learning method on the task of terrain segmentation for a dataset of images collected by the Mars Curiosity rover.

In summary, the contributions of this work are:
\begin{itemize}
    \item A deep learning framework to perform image-based terrain segmentation with contrastive pretraining
    \item Performance analysis of terrain segmentation with small sets of labeled images
    \item Analysis and visualization of transferability of generalized image features obtained through large-scale contrastive pretraining
\end{itemize}

\section{Related Work}
\label{sec:literature}

\subsection{Semantic Segmentation in images}
Semantic segmentation can be defined as the task of assigning class labels to individual pixels of an input image. Modern methods for image semantic segmentation mostly rely on deep convolutional neural networks (CNN), and some examples of popular network architectures for this are SegNet \cite{segnet2015}, U-Net \cite{unet2015}, and DeepLab \cite{deeplab2017}. The main difference between CNNs for classification and CNNs for semantic segmentation is that CNNs for classification predict a single output label for a given image, whereas CNNs for segmentation predict multiple output labels (i.e., one for each pixel) for a given image. As such, segmentation networks usually have specialized layers (e.g., skip connections for the case of U-Net \cite{unet2015} and atrous convolutions for the case of DeepLab \cite{deeplab2017}) to enable the network layers to capture sufficient spatial information to predict a dense output.

\subsection{Terrain segmentation}
Terrain segmentation is a subset of semantic segmentation where the pixels of an input image are labeled with different types of terrain features such as sand, soil, and rock. Terrain segmentation is especially important for robotics due to its applications in obstacle avoidance, traversability estimation, sample collection, and energy-aware path planning \cite{gonzalez2018, dimastrogiovanni2020}. The basic approach to terrain segmentation is to define a basic set of terrain classes and apply a standard machine learning algorithm such as Support Vector Machines (SVM) \cite{dimastrogiovanni2020} or Convolutional Neural Networks (CNN) \cite{gonzalez2018}. The segmentation results can be further refined by applying Fast Fourier Transforms (FFT) \cite{bai2019} or connected component labelling \cite{miliaresis2004}. TU-Net and TDeepLab have been proposed to allow terrain segmentation to be more robust to illumination changes \cite{tunet2019}, but they rely on data fusion with thermal images, which is not always available. 
In the domain of lunar images, terrain segmentation has been carried out using wavelet decomposition \cite{liang2017lunar} and histogram of features \cite{jiang2015}. However, these studies only use orbital images from the Chang'e mission and not the surface images. 
On the other hand, an initial study of Mars terrain segmentation is carried out by the Soil Property and Object Classification (SPOC) project \cite{spoc2016}, which deals with terrain classification for the purpose of landing site traversability analysis and slip prediction for the Mars Science Laboratory (MSL) mission. This study is extended through the MAARS (Machine learning-based Analytics for Automated Rover Systems) effort \cite{maars2020}, which aims to further enhance  the  productivity and safety of planetary rover missions. The MAARS effort includes automated scientific captioning of terrain images for the purpose of image similarity search and retrieval and resource-aware path planning for the Mars rover that combines vision-based terrain segmentation with a terramechanics model. However, these existing studies for terrain segmentation still rely on standard supervised learning techniques which require a large amount of time and manual effort to label the training images. Some research studies have attempted to resolve the issue of lack of training data, at least for Earth terrain, by using self-supervision techniques where data from sensors from other modalities can be used to provide supervision for data from the vision sensor. This can come in the form of using acoustic sensors to learn vehicle-terrain interaction sounds \cite{zurn2021} or using a combination of motion sensors and Lidar to learn the traversability of different types of terrain features \cite{kahn2020}. However, these techniques may perform poorly for the case of planetary rovers since physical interactions between the rover and the environment are expensive and occur at a slower rate.

\subsection{Transfer learning for images}
Transfer learning refers to the process where a model which is initially trained on one machine learning task is reused or altered for a different machine learning task. The main benefit of transfer learning is data-efficiency, since it allows a model to make use of information from multiple datasets and tasks and this is advantageous when few labels are available for the actual task (e.g. in one-shot learning or few-shot learning). A common approach for training a deep neural network is to pretrain it on a standard image classification task with a large dataset such as ImageNet and then finetune it with a smaller dataset for the specific application. This approach can be applied for the case of off-road driving by pretraining on on-road driving images \cite{holder2016}. This approach has also been used with some success on classification of the Mars Science Laboratory (MSL) surface images and High Resolution Imaging Science Experiment (HiRISE) orbital images \cite{wagstaff2018}, where pretraining is performed using the ImageNet dataset. However, this approach may suffer from domain shift due to the differences in image properties between the types of images used for pretraining and the types of images used for finetuning \cite{ltf2020}. Another approach for training deep neural networks, especially for robotics problems, is to utilize simulations to generate synthetic training data \cite{sharma2019}. Synthetic generation of data is advantageous in that a large number of images can be easily and procedurally generated; however, it can also suffer from domain shift issues when the images in the source domain (i.e., simulations) differ significantly from images in the target domain (i.e., real-world) in terms of object types, appearance, resolution, camera field-of-view, and lighting.

Domain adaptation is a large field of research within transfer learning that deals with ways to help a model trained on a source distribution to adapt to a different target distribution. Methods such as subspace alignment \cite{fernando2013} and flow kernels \cite{gong2012} have been proposed to determine a mapping function that can align properties of images in the source domain with the target one. On the other hand, methods such as network adaptation \cite{ganin2015} and correlation alignment \cite{coral2016} enable deep neural networks to better generalize to the target domain based on introducing additional loss functions such as domain classification loss and regularization loss. Later works attempt to improve the generalizability of pixel-level prediction tasks by constructing pivot information that is common knowledge shared across domains \cite{ouali2020} or by enforcing consistency in perturbed images \cite{lv2020}. These methods can be considered unsupervised in that they only use labels in the source domain and do not require labels in the target domain. Thus, these methods are suitable for applications where labels in the target domain are expensive and time-consuming to obtain, which is the case for terrain segmentation for planetary rovers. However, domain adaptation does not completely solve other issues in transfer learning between datasets such as taxonomy inconsistencies and inconsistencies in segmentation detail \cite{lambert2020}.

\subsection{Representation learning for images}
A closely-related idea is representation learning, where the focus is not on training a model to predict specific outputs such as classification or segmentation but on training a general model that can extract useful features from high-dimensional data \cite{cpc2019}. Representation learning usually involves setting up an auxiliary task where unlabeled samples can still be used to generate weak supervision for training a neural network. In the case of associative embeddings \cite{newell2017}, this involves predicting a vector embedding for each pixel that represents its group assignment in the image. In the case of SimCLR \cite{simclr2020}, this involves applying data augmentations such as cropping, distorting, and blurring, and training the network to recognize matching images and mismatched images in the presence of these augmentations with a contrastive loss function. In the case of DINO \cite{dino2021}, this involves training a student model, which has access to local views of an image, to match the prediction of a teacher model, which has access to global views of an image, with a cross-entropy loss function.

In this research, we propose to apply transfer learning to the task of terrain segmentation in the form of contrastive pretraining. Existing work such as SimCLR\cite{simclr2020} and CPC\cite{cpc2019} mostly consider the classification task, and not the segmentation task, whereas in the proposed research, we make use of contrastive loss to learn a rich representation of image features from a neural network and specifically fine-tune the network for the terrain segmentation task.

%===============================================================================

\section{Methodology}
\label{sec:methodology}

\subsection{Problem description}

This work leverages the first large-scale dataset of Mars images labeled for the purposes of terrain classification and traversability assessment --- AI4Mars \cite{swan2021}.
This dataset consists of images from Curiosity rover's navigation camera (NAVCAM) and color mast camera (Mastcam). 
The training set consists of 16,064 images whose labels were obtained using a novel crowdsourcing approach. In this approach, citizen scientists voluntarily identified four classes --- soil, bedrock, sand, and big rock --- in MSL images. These submissions were then aggregated based on several pixel-level agreement heuristics described in \cite{swan2021}. The "gold standard" test set includes 322 images where each pixel's label was unanimously determined by three experts. 

In addition to the four classes, the AI4Mars images also include a null class (i.e., unlabeled regions of the image), masks for portions of the Curiosity rover that appear in the images, and masks for portions of the images that are beyond 30 m from the rover.
These masks are treated as separate classes (``rover" and ``background", respectively) in the training process, while the predictions associated with unlabeled pixels are dropped from the calculation of training and evaluation performance metrics.
Thus, unless otherwise specified, this work aims to segment MSL images into six classes.

Minimal preprocessing is performed on the images in the training pipeline; images and masks are resized to 512x512 with nearest neighbor interpolation from the original 1024x1024. Additionally, images are normalized to a range of 0-1 and uniformly brightened by 50\% (i.e., multiplying all pixels by 1.5 and clipping the image to a range of 0-1).

\begin{figure*}
    \centering
    \includegraphics[width=1\linewidth]{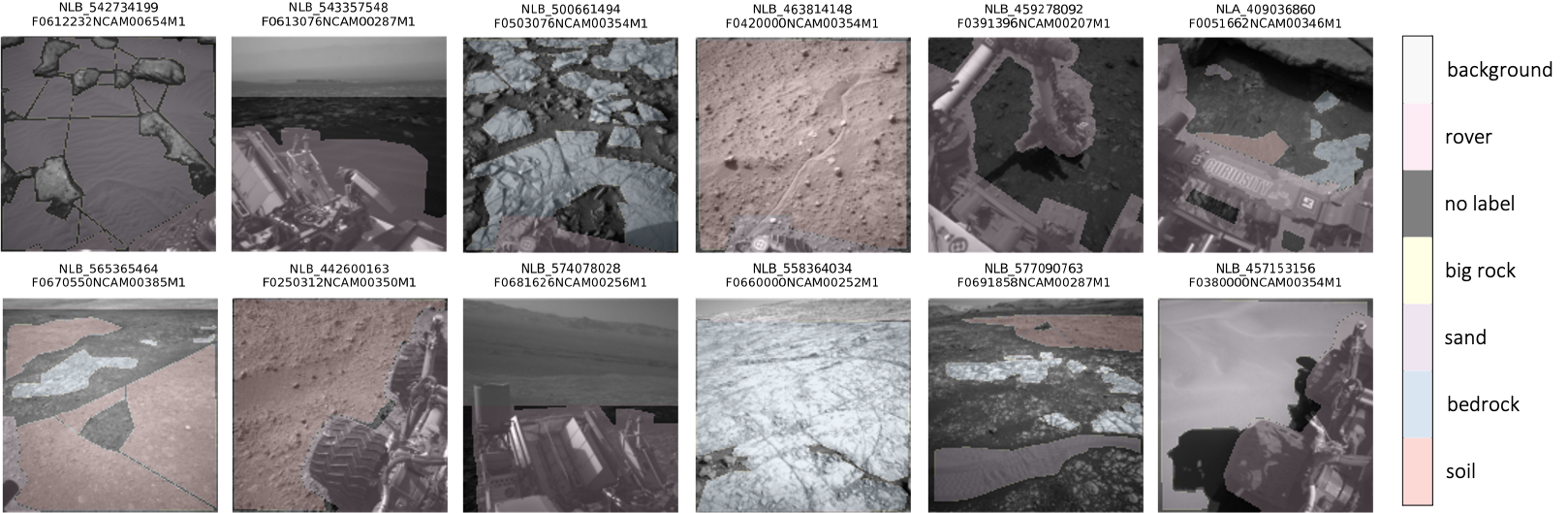} 
    \caption[]{Sample MSL images and labels from the AI4Mars dataset \cite{swan2021}.} 
    \label{fig:sample_data}
\end{figure*}

\subsection{Network architecture}

The SimCLR framework consists of three main components:

\begin{enumerate}
    \item A large task-agnostic backbone/encoder, $f(\cdot)$, that learns generalized visual embeddings through self-supervised pretraining
    \item An MLP projection head, $g(\cdot)$, that maps the latent embedding space to the one where contrastive loss is applied during self-supervised pretraining
    \item A smaller, task-specific head that is attached either directly to the base encoder $f(\cdot)$ as in \cite{simclr2020}, or to a middle layer of the projection head $g(\cdot)$ as in \cite{simclrv2}. This task-specific head is used in a supervised finetuning task with presumably limited labels.
\end{enumerate}

This research uses the residual network (ResNet) \cite{he2015} as the base encoder. Two main modifications on a ``vanilla'' ResNet are incorporated from \cite{simclrv2} --- a 2x wider (i.e., twice the filters in each layer) ResNet and selective kernels (SK) \cite{sk2019}. Deeper and wider ResNets were shown to benefit a semi-supervised learning regime more than supervised learning approaches \cite{simclr2020,simclrv2}, while SK reduces model complexity (thus increasing parameter efficiency) by allowing neurons to dynamically adjust their receptive field sizes. 

With this task addressing a segmentation problem, atrous convolutions \cite{liang2017} are used for the task-specific output head, as opposed to the linear classifiers used in \cite{simclrv2}. The combination of a ResNet encoder and atrous convolutional decoder has previously been proposed for the DeepLab architecture \cite{deeplab2017}, except that in this work, we also incorporate self-supervised learning modifications to the ResNet encoder.

\subsection{Training procedure}
\subsubsection{Self-supervised pretraining}
SimCLR is a self-supervised contrastive learning approach where an encoder is trained to recognize that two augmented views $i$ and $j$ of the same image should have similar embeddings  $z_i$ and $z_j$ in the latent space. This generalized representation learning is achieved by training the network to minimize a contrastive loss $l_{i,j}$, given as:
\begin{subequations}
\begin{align}
    \label{eq:contrastive_loss}
    &l_{i,j} = -\log\frac{S_{i,j}}{\sum_{k=1}^{2N}\mathbbm{1}_{[k\neq i]}S_{i,k}} \\
    &S_{i,j} = \exp{\left(\text{sim}(z_i, z_j)/\tau\right)} \\
    &\text{sim}(z_i, z_j) = \frac{z_i^\top z_j }{ \left\lVert z_i \right\rVert \left\lVert z_j \right\rVert }
\end{align}
\end{subequations}

where $S$ captures the temperature-scaled similarity between two vector embeddings in latent space, 
$\text{sim}$ is the cosine similarity between the two vectors, 
$\tau$ is a temperature scalar, 
and $N$ is the total number of images in the batch.

\begin{figure}
    \centering
    \includegraphics[width=0.95\columnwidth]{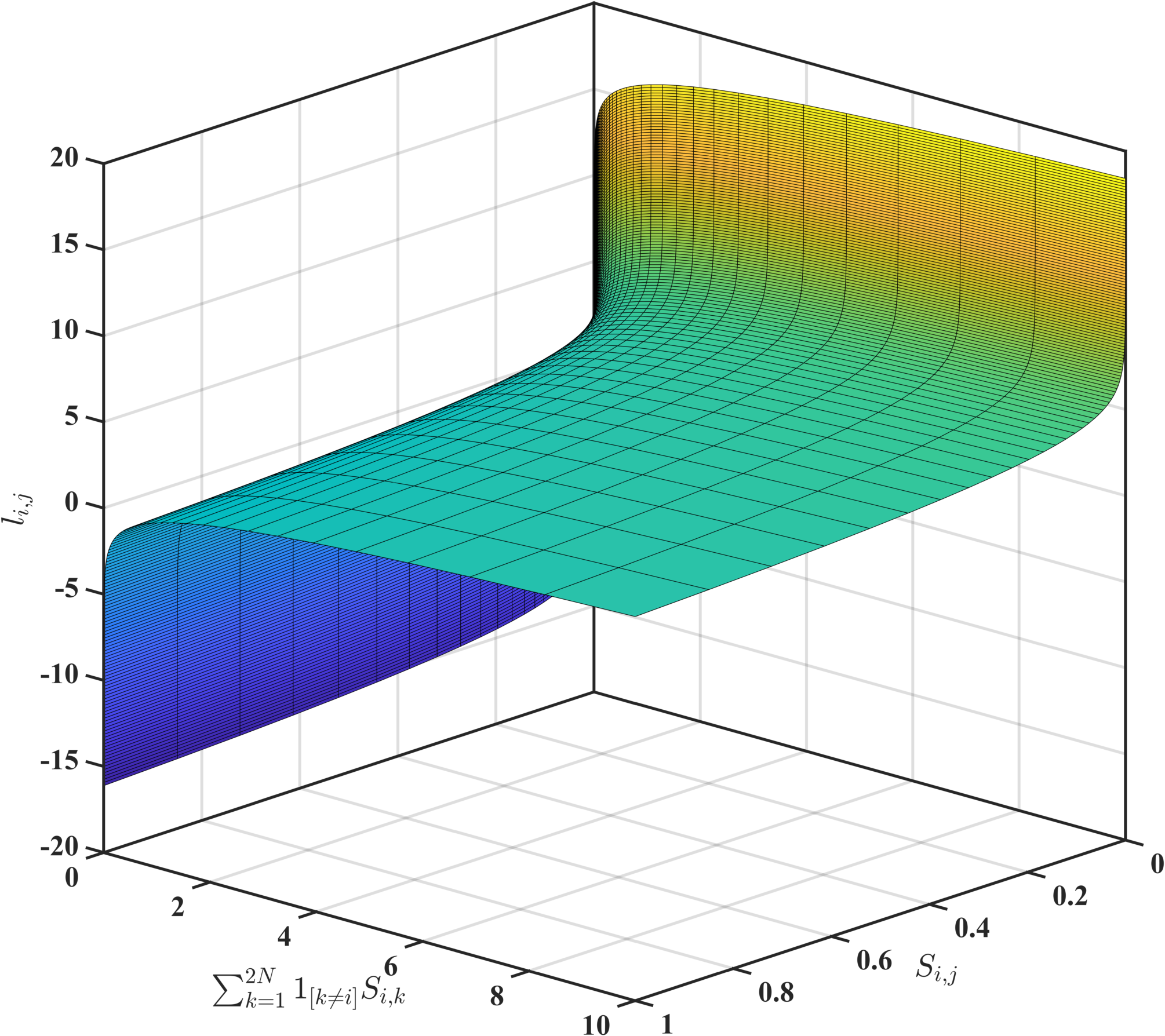}   
    \caption[]{Representative loss surface for the contrastive loss used in self-supervised pretraining} 
    \label{fig:contrastive_loss}
\end{figure}

In Eq. \ref{eq:contrastive_loss}, the normalized cross-entropy loss decreases as $S_{i,j}$ increases to 1 and $S_{i,k}$ decreases. In other words, loss decreases as the network learns to generate similar embeddings from augmented views of the same image and distant embeddings for augmented views from different images. A notional loss surface is shown in Fig. \ref{fig:contrastive_loss}.

It has been found that contrastive learning benefits from larger batch sizes compared to supervised learning. The benefit stems from the increased number of negative examples ($k$ in Eq. \ref{eq:contrastive_loss}) per positive pair of augmented views. For instance, a batch size of up to 8192 was used in \cite{simclr2020} and a batch size of 4096 was used in \cite{simclrv2} for contrastive pretraining on the ImageNet ILSVRC-2012 dataset. As such, while contrastive learning offers the benefit of training on unlabeled data, this benefit comes at the expense of increased computational capacity required for the larger batch sizes and longer training durations. 

Because of the computational cost required for contrastive pretraining, this task uses the task-agnostic ResNet-50 (2x width) weights from \cite{simclrv2} that have been optimized to learn generalized visual representations without performing further contrastive learning. The pretrained weights are incorporated directly into the supervised finetuning stage with MSL images, as described in the following section.

\subsubsection{Supervised Finetuning}

For supervised finetuning, a segmentation head consisting of an atrous convolution block and a resize layer is attached to a ResNet-50 base encoder. The atrous block performs convolutions on the encoder output at three different dilation rates (6, 12, and 18), and each convolutional layer learns 256 filters with complementary (effective) fields of view. These filters are then concatenated and resized to the original image size before being passed through a 1x1 convolutional layer to obtain the final segmentation logits. With the AI4Mars dataset used in the current work, the dimensions of the outputs are (512, 512, 6). The combination of a 2x ResNet-50 and an atrous block with 256 filters per layer results in a network with 171,172,160 parameters.

A pixel-wise cross entropy loss function which neglects unlabeled pixels is used in this work. Weighting the loss function by relative pixel frequencies was found in \cite{swan2021} to offer little in the way of performance but incur additional computational cost, and was therefore not investigated. 
A batch size of 8 and a learning rate of 0.057 $\left(0.002 \times \sqrt{8}\right)$ is used for finetuning. While the specified number of epochs is 50 for each experiment, an early stopping callback terminates the runs upon convergence.
Each experiment is run on a single compute node with four Nvidia V100 GPUs.

\section{Results}
\label{sec:result}

\subsection{Analysis of contrastive pretraining performance}

\begin{figure}[H]
    \centering
    \includegraphics[width=0.8\columnwidth]{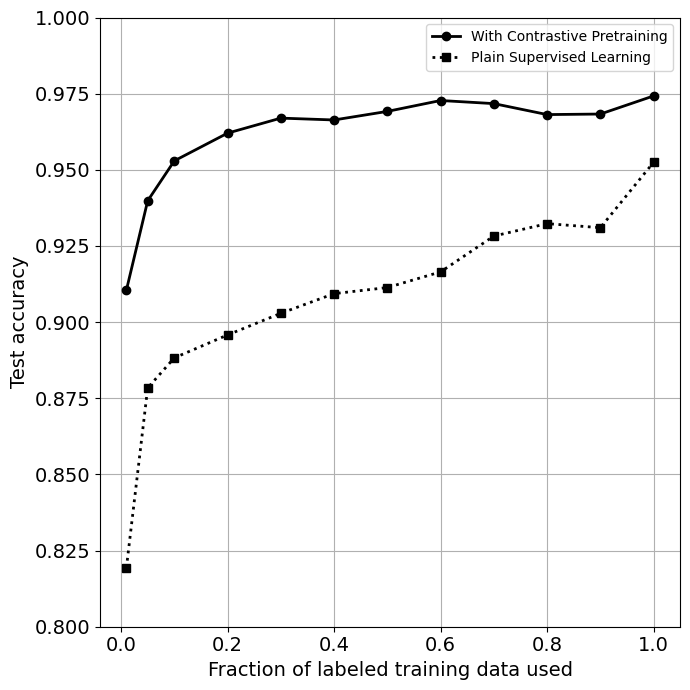}
    \caption[]{Graph of test accuracy vs fraction of labeled training data used. The total number of labeled training images is 16,064.} 
    \label{fig:acc_vs_labelfraction}
\end{figure}

Figure \ref{fig:acc_vs_labelfraction} shows a comparison between the test accuracy of the proposed contrastive pretraining method for terrain segmentation, where the network weights are initialized using a contrastive loss function, versus plain supervised learning, where the network weights are initialized randomly. The number of training images used ranges from 161 (1\% of the original dataset) to 16,064 (100\% of the original training set). The graph shows that the proposed contrastive pretraining method has a clear advantage over plain supervised learning in segmenting Mars terrain images. 
In particular, the proposed method consistently outperforms the supervised learning model, with the performance margin widening as the fraction of labeled training data decreases. This shows that contrastive pretraining is especially helpful in applications where the number of labeled training samples available is small.

\begin{figure*}[!htbp]
    \centering
    \begin{subfigure}{0.4\textwidth}
        \includegraphics[width=1\linewidth]{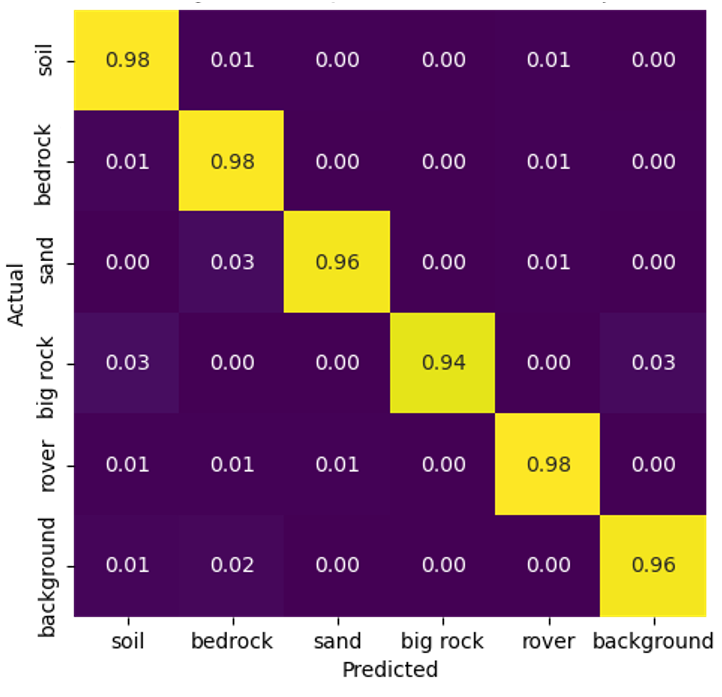}
        \caption[]{(Pretrained) 100\%; Accuracy: 97.4\%}
        \label{fig:pretrained_100pct}
    \end{subfigure}
    \begin{subfigure}{0.4\textwidth}
        \includegraphics[width=1\linewidth]{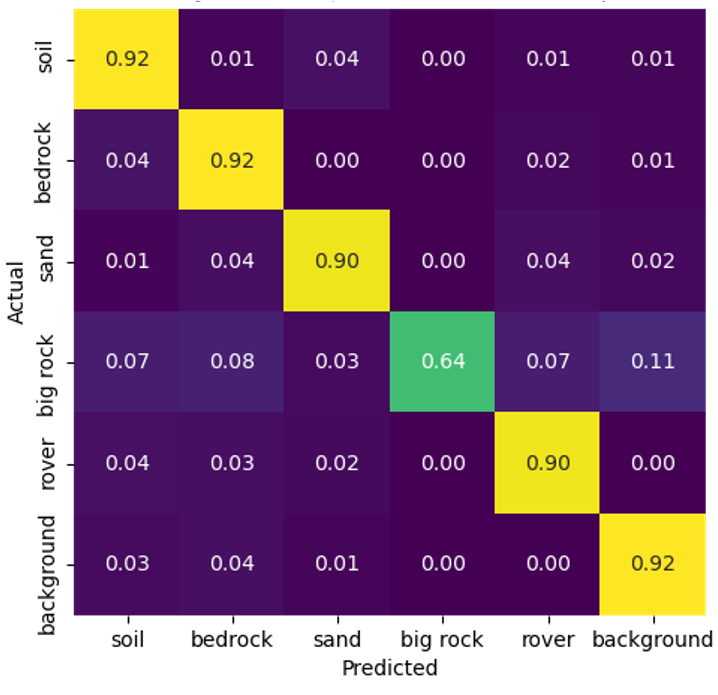}
        \caption[]{(Pretrained) 1\%; Accuracy: 91.1\%}
        \label{fig:pretrained_1pct}
    \end{subfigure}
    \caption{Contrastive pretrained ResNet-50 (2x width) finetuned using 100\% and 1\% of training samples. Confusion matrices are normalized with respect to the total number of true samples, i.e., columns in each row sum up to 100\%.}
    \label{fig:pretrained_comparison}
\end{figure*}

\begin{figure}[ht]
    \centering
    \includegraphics[width=1\columnwidth]{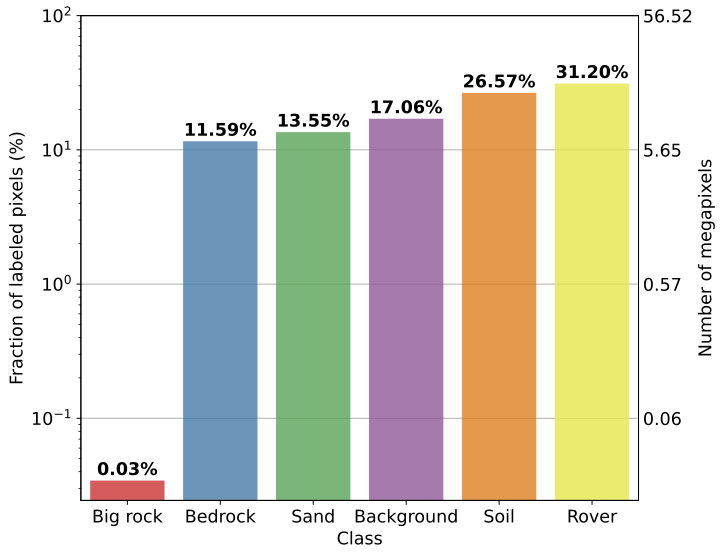}    
    \caption[]{Distribution of pixel counts across different classes. Only labeled pixels are included in this figure; pixels labeled ``NULL'' in the AI4Mars dataset are ignored.} 
    \label{fig:test_distribution}
\end{figure}

The confusion matrices in Figures \ref{fig:pretrained_comparison} and \ref{fig:supervised_comparison} show the per-class recall values for the pretrained and supervised cases, respectively, with 1\% and 100\% training sets. Model performance on the ``Big Rock'' class is lower than the other classes across all runs. This is caused by the imbalance in class frequency as shown in Figure \ref{fig:test_distribution}, where ``Big Rock'' has orders of magnitude fewer pixels compared to the other classes.
Indeed, the ``Big Rock'' class is only present in two images in the test set.
Contrastive pretraining increases recall by 6\% in both scenarios (1\% and 100\% data), but exhibits more consistent performance across the entire spectrum of training data used.

\begin{figure}[ht]
    \centering
    \includegraphics[width=0.9\linewidth]{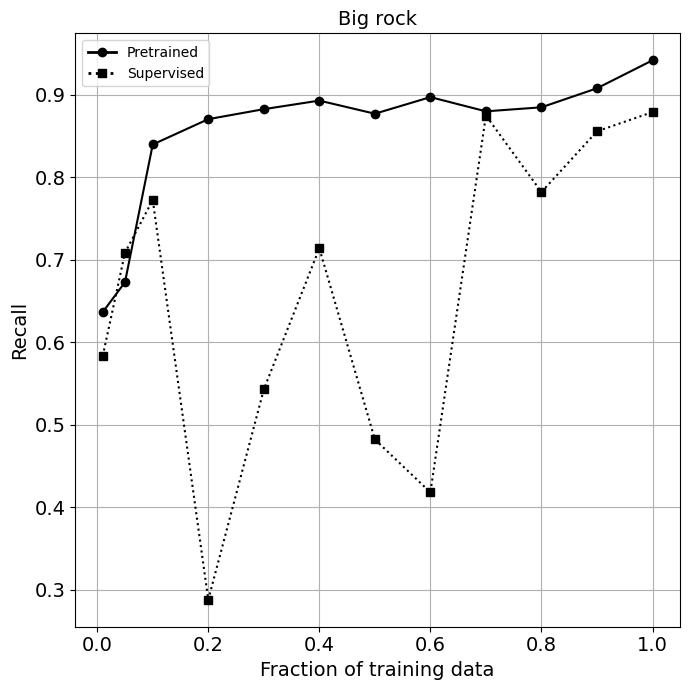}   
    \caption[]{Comparison between contrastive pretraining and plain supervised learning performance on the ``Big Rock'' class as the provided training data is varied.} 
    \label{fig:big_rock_comparison}
\end{figure}

Figure \ref{fig:big_rock_comparison} shows a comparison between contrastive pretraining and plain supervised learning on rare objects (i.e. the Big Rock class). The contrastive pretrained model consistently achieves around 90\% recall with training sets larger than 20\% (around 3200 images), whereas the baseline model shows significant variance in its performance. This trend indicates the potential for contrastive and other self-supervised learning techniques to address class imbalance and the presence of rare objects. However, Fig. \ref{fig:big_rock_comparison} also shows that the current models are unable to achieve reasonable recall rates when the training set is smaller than 20\% (around 3200 images). Some possible methods to address this issue include finetuning on larger pretrained models (for instance, the largest model used in \cite{simclrv2} is a 3x ResNet-152 with 795M parameters), different model architectures (e.g., vision transformers), and large-scale contrastive pretraining on domain-specific datasets.

\subsection{Visualization of terrain segmentation results}
\begin{figure*}[!htbp]
    \centering
    \begin{subfigure}{0.4\textwidth}
        \includegraphics[width=1\linewidth]{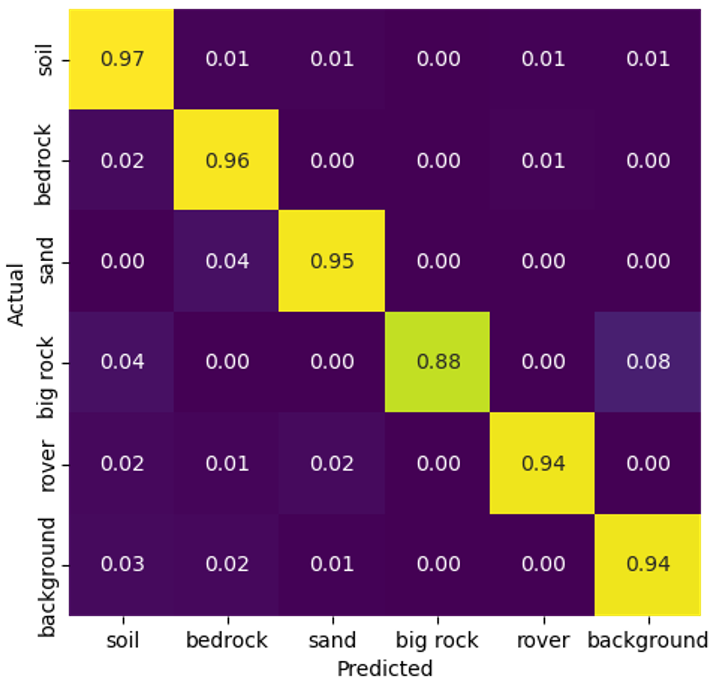}
        \caption[]{(Supervised) 100\%; Accuracy: 95.3\%}
        \label{fig:supervised_100pct}
    \end{subfigure}
    \begin{subfigure}{0.4\textwidth}
        \includegraphics[width=1\linewidth]{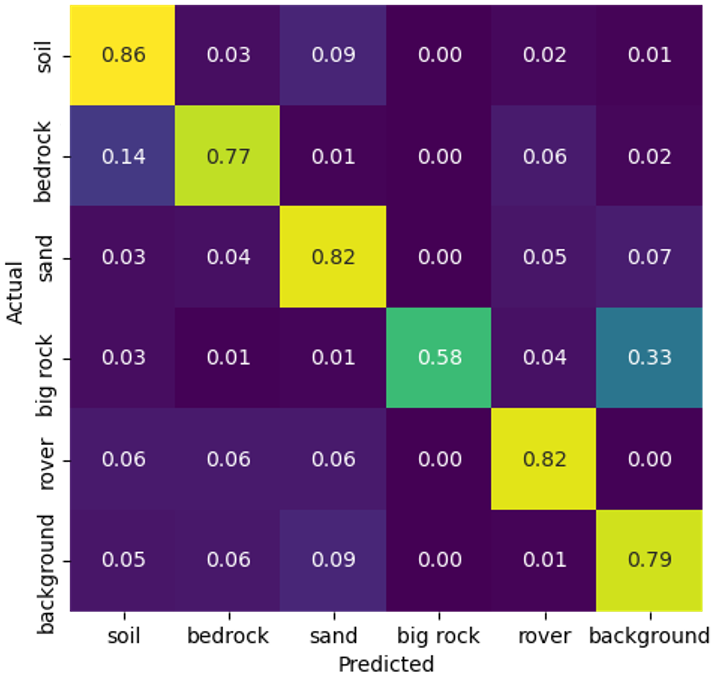}
        \caption[]{(Supervised) 1\%; Accuracy: 81.9\%}
        \label{fig:supervised_1pct}
    \end{subfigure}
    \caption{Plain supervised ResNet-50 (2x width) trained using 100\% and 1\% of training samples. Confusion matrices are normalized with respect to the total number of true samples, i.e., columns in each row sum up to 100\%,}
    \label{fig:supervised_comparison}
\end{figure*}

\begin{figure*}
    \centering
    \includegraphics[width=1\linewidth]{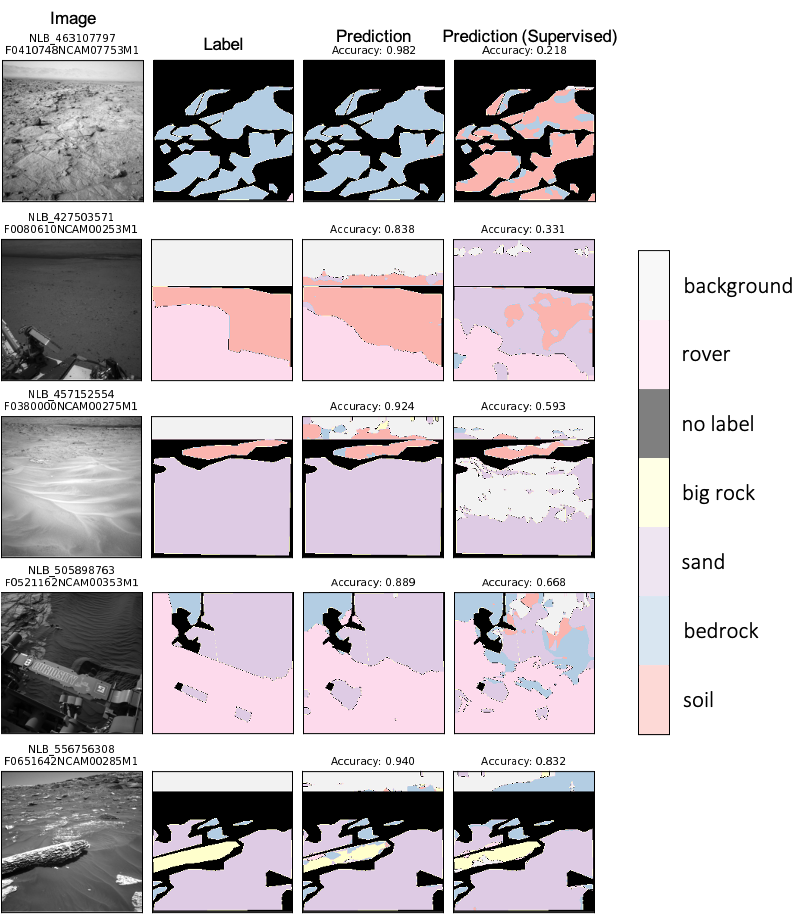} 
    \caption[]{Sample predictions where the contrastive pretrained model outperforms the plain supervised model. Results are shown for the 1\% training set case, with both models converging at 42 epochs.} 
    \label{fig:pred_1pct_pretrained_better}
\end{figure*}

\begin{figure*}
    \centering
    \includegraphics[width=1\linewidth]{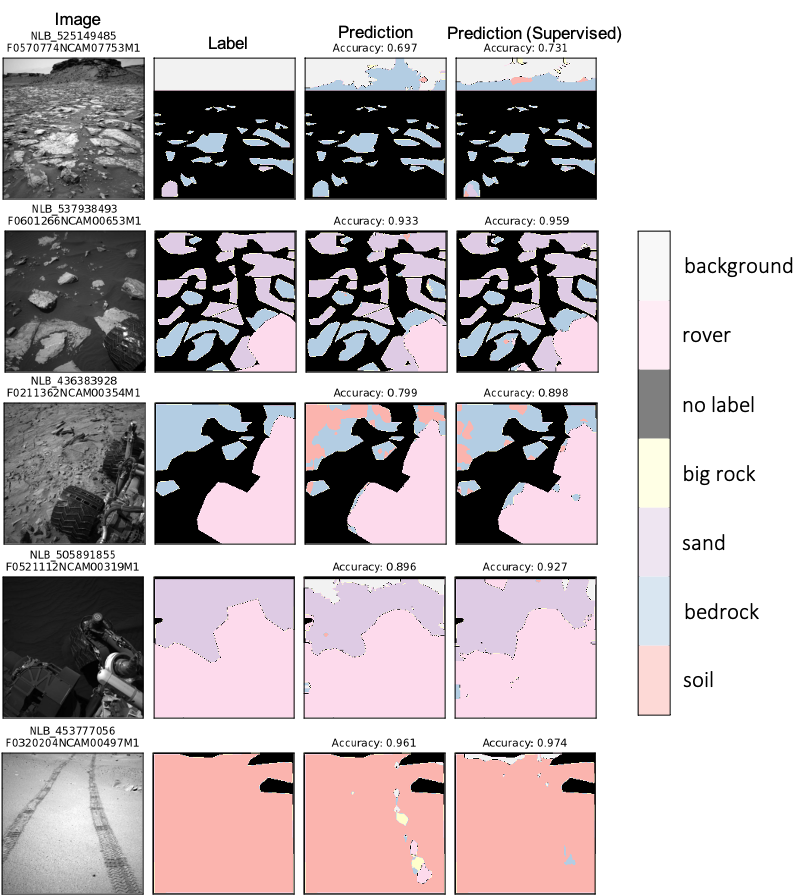} 
    \caption[]{Sample predictions where the plain supervised model outperforms the contrastive pretrained. Results are shown for the 1\% training set case, with both models converging at 42 epochs.} 
    \label{fig:pred_1pct_supervised_better}
\end{figure*}

Figures \ref{fig:pred_1pct_pretrained_better} and \ref{fig:pred_1pct_supervised_better} show the segmentation masks predicted by the contrastive pretrained model compared to the plain supervised model when 1\% of the training data is used. Under this limited labels scenario, the pretrained model can better discriminate between similar classes such as ``bedrock'', ``soil'', and ``sand''.
This is shown through the top two examples in Fig. \ref{fig:pred_1pct_pretrained_better}, where the contrastive pretrained model separates the classes correctly whereas the plain supervised model exhibits significant error in the three aforementioned classes. The observations in the segmentation masks for the plain supervised model are consistent with the confusion matrix in Fig. \ref{fig:supervised_1pct}, which highlights significant ``confusion'' between soil and bedrock, as well as between rock and background.

Figure \ref{fig:pred_1pct_supervised_better} shows certain cases where the plain supervised model performed better than the contrastive pretrained model. For the occasional scenarios where the supervised model performed better, the contrastive pretrained model did not exhibit major drops in performance, with accuracy differences ranging from only 1\% to 10\%. Another notable observation is that the crowdsourced ``ground truth'' label may not completely capture nuances in the image. For example, the ground truth label in Figure \ref{fig:pred_1pct_pretrained_better} Row 2 does not accurately represent the (diagonal) outline of the rover, whereas the predicted output from our contrastive pretrained model captures the rover outline at a fine level of detail. In addition, the tire tracks in Figure \ref{fig:pred_1pct_supervised_better} Row 5 are not labeled in the ground truth, but they are detected in the predicted output from our contrastive pretrained model.

% Visualization of segmentation results

\section{Conclusion}
\label{sec:conclusion}

In conclusion, this research proposes an image-based terrain segmentation method using contrastive pretraining techniques. Our potential relevance to future planetary science missions is the development of a weakly-supervised, data-efficient deep learning framework that would enable and enhance future scientific investigations on planetary data by facilitating the data-driven analysis and interpretation of planetary image-based datasets. The model can be applied at a large-scale to new images and compute high-dimensional feature-rich representations that are useful to multiple downstream (image) analysis tasks. The proposed method has potential wider-scale contributions in terms of improving the robustness and generalizability of image segmentation, as well as enabling accurate terrain segmentation for improved navigation, safety, and search capabilities of mobile robots. One area of future work would be to extend the proposed method from the task of segmenting surface images to the task of segmenting images from aerial and/or orbital platforms. This task is relevant to the planning of future planetary science missions, and their applications include (but are not limited to) improved landing site selection for crew safety as well as enabling remote sensing identification of target destinations with high scientific value (e.g., identifying locations similar to existing images with potential biosignatures).

%%%%%%%%%%%%%%%%%%%%%%%%%%%%%%%%%%%%%%%%%%%%%%%%%%%%%%%%%%%%%%%%%%%%%%%%%%%%%%%%%%%%%%%%%%%%%%%%%%%%%%
\section*{Acknowledgments}
A portion of this research was carried out at the Jet Propulsion Laboratory, California Institute of Technology, under a contract with the National Aeronautics and Space Administration (80NM0018D0004). The authors would like to thank Hiro Ono and Michael Swan for facilitating access to AI4Mars. U.S. Government sponsorship acknowledged. The authors acknowledge the Texas Advanced Computing Center (TACC) at The University of Texas at Austin for providing HPC resources that have contributed to the research results reported herein. URL: http://www.tacc.utexas.edu

%%%%%%%%%%%%%%%%%%%%%%%%%%%%%%%%%%%%%%%%%%%%%%%%%%%%%%%%%%%%%%%%%%%%%%%%%%%%%%%%%%%%%%%%%%%%%%%%%%%%%%
\bibliographystyle{IEEEtran}
\bibliography{references.bib}

\begin{thebibliography}{10}
\providecommand{\url}[1]{#1}
\csname url@samestyle\endcsname
\providecommand{\newblock}{\relax}
\providecommand{\bibinfo}[2]{#2}
\providecommand{\BIBentrySTDinterwordspacing}{\spaceskip=0pt\relax}
\providecommand{\BIBentryALTinterwordstretchfactor}{4}
\providecommand{\BIBentryALTinterwordspacing}{\spaceskip=\fontdimen2\font plus
\BIBentryALTinterwordstretchfactor\fontdimen3\font minus
  \fontdimen4\font\relax}
\providecommand{\BIBforeignlanguage}[2]{{%
\expandafter\ifx\csname l@#1\endcsname\relax
\typeout{** WARNING: IEEEtran.bst: No hyphenation pattern has been}%
\typeout{** loaded for the language `#1'. Using the pattern for}%
\typeout{** the default language instead.}%
\else
\language=\csname l@#1\endcsname
\fi
#2}}
\providecommand{\BIBdecl}{\relax}
\BIBdecl

\bibitem{williford2018}
\BIBentryALTinterwordspacing
K.~H. Williford, K.~A. Farley, K.~M. Stack, A.~C. Allwood, D.~Beaty, L.~W.
  Beegle, R.~Bhartia, A.~J. Brown, M.~{de la Torre Juarez}, S.-E. Hamran, M.~H.
  Hecht, J.~A. Hurowitz, J.~A. Rodriguez-Manfredi, S.~Maurice, S.~Milkovich,
  and R.~C. Wiens, ``Chapter 11 - the nasa mars 2020 rover mission and the
  search for extraterrestrial life,'' in \emph{From Habitability to Life on
  Mars}, N.~A. Cabrol and E.~A. Grin, Eds.\hskip 1em plus 0.5em minus
  0.4em\relax Elsevier, 2018, pp. 275--308. [Online]. Available:
  \url{https://www.sciencedirect.com/science/article/pii/B9780128099353000104}
\BIBentrySTDinterwordspacing

\bibitem{colaprete2019}
A.~{Colaprete}, D.~{Andrews}, W.~{Bluethmann}, R.~C. {Elphic}, B.~{Bussey},
  J.~{Trimble}, K.~{Zacny}, and J.~E. {Captain}, ``{An Overview of the
  Volatiles Investigating Polar Exploration Rover (VIPER) Mission},'' in
  \emph{AGU Fall Meeting Abstracts}, vol. 2019, Dec. 2019, pp. P34B--03.

\bibitem{lu2009}
L.~Lu, C.~Ordonez, E.~G. Collins, and E.~M. DuPont, ``Terrain surface
  classification for autonomous ground vehicles using a 2d laser stripe-based
  structured light sensor,'' in \emph{2009 IEEE/RSJ International Conference on
  Intelligent Robots and Systems}, 2009, pp. 2174--2181.

\bibitem{maars2020}
M.~Ono, B.~Rothrock, K.~Otsu, S.~Higa, Y.~Iwashita, A.~Didier, T.~Islam,
  C.~Laporte, V.~Sun, K.~Stack, J.~Sawoniewicz, S.~Daftry, V.~Timmaraju,
  S.~Sahnoune, C.~A. Mattmann, O.~Lamarre, S.~Ghosh, D.~Qiu, S.~Nomura, H.~Roy,
  H.~Sarabu, G.~Hedrick, L.~Folsom, S.~Suehr, and H.~Park, ``Maars: Machine
  learning-based analytics for automated rover systems,'' in \emph{2020 IEEE
  Aerospace Conference}, 2020, pp. 1--17.

\bibitem{spoc2016}
\BIBentryALTinterwordspacing
B.~Rothrock, R.~Kennedy, C.~Cunningham, J.~Papon, M.~Heverly, and M.~Ono,
  \emph{SPOC: Deep Learning-based Terrain Classification for Mars Rover
  Missions}. [Online]. Available:
  \url{https://arc.aiaa.org/doi/abs/10.2514/6.2016-5539}
\BIBentrySTDinterwordspacing

\bibitem{tunet2019}
Y.~Iwashita, K.~Nakashima, A.~Stoica, and R.~Kurazume, ``Tu-net and tdeeplab:
  Deep learning-based terrain classification robust to illumination changes,
  combining visible and thermal imagery,'' in \emph{2019 IEEE Conference on
  Multimedia Information Processing and Retrieval (MIPR)}, 2019, pp. 280--285.

\bibitem{ltf2020}
\BIBentryALTinterwordspacing
J.~Guo, M.~Gong, T.~Liu, K.~Zhang, and D.~Tao, ``{LTF}: A label transformation
  framework for correcting label shift,'' in \emph{Proceedings of the 37th
  International Conference on Machine Learning}, ser. Proceedings of Machine
  Learning Research, H.~D. III and A.~Singh, Eds., vol. 119.\hskip 1em plus
  0.5em minus 0.4em\relax PMLR, 13--18 Jul 2020, pp. 3843--3853. [Online].
  Available: \url{https://proceedings.mlr.press/v119/guo20d.html}
\BIBentrySTDinterwordspacing

\bibitem{simclr2020}
\BIBentryALTinterwordspacing
T.~Chen, S.~Kornblith, M.~Norouzi, and G.~Hinton, ``A simple framework for
  contrastive learning of visual representations,'' in \emph{Proceedings of the
  37th International Conference on Machine Learning}, ser. Proceedings of
  Machine Learning Research, H.~D. III and A.~Singh, Eds., vol. 119.\hskip 1em
  plus 0.5em minus 0.4em\relax PMLR, 13--18 Jul 2020, pp. 1597--1607. [Online].
  Available: \url{http://proceedings.mlr.press/v119/chen20j.html}
\BIBentrySTDinterwordspacing

\bibitem{simclrv2}
T.~Chen, S.~Kornblith, K.~Swersky, M.~Norouzi, and G.~E. Hinton, ``Big
  self-supervised models are strong semi-supervised learners,'' \emph{Advances
  in Neural Information Processing Systems}, vol.~33, pp. 22\,243--22\,255,
  2020.

\bibitem{cpc2019}
A.~van~den Oord, Y.~Li, and O.~Vinyals, ``Representation learning with
  contrastive predictive coding,'' 2019.

\bibitem{dino2021}
M.~Caron, H.~Touvron, I.~Misra, H.~J\'egou, J.~Mairal, P.~Bojanowski, and
  A.~Joulin, ``Emerging properties in self-supervised vision transformers,'' in
  \emph{Proceedings of the International Conference on Computer Vision (ICCV)},
  2021.

\bibitem{segnet2015}
\BIBentryALTinterwordspacing
V.~Badrinarayanan, A.~Kendall, and R.~Cipolla, ``Segnet: A deep convolutional
  encoder-decoder architecture for image segmentation.'' \emph{CoRR}, vol.
  abs/1511.00561, 2015. [Online]. Available:
  \url{http://dblp.uni-trier.de/db/journals/corr/corr1511.html#BadrinarayananK15}
\BIBentrySTDinterwordspacing

\bibitem{unet2015}
\BIBentryALTinterwordspacing
O.~Ronneberger, P.Fischer, and T.~Brox, ``U-net: Convolutional networks for
  biomedical image segmentation,'' in \emph{Medical Image Computing and
  Computer-Assisted Intervention (MICCAI)}, ser. LNCS, vol. 9351.\hskip 1em
  plus 0.5em minus 0.4em\relax Springer, 2015, pp. 234--241, (available on
  arXiv:1505.04597 [cs.CV]). [Online]. Available:
  \url{http://lmb.informatik.uni-freiburg.de/Publications/2015/RFB15a}
\BIBentrySTDinterwordspacing

\bibitem{deeplab2017}
L.-C. Chen, G.~Papandreou, I.~Kokkinos, K.~Murphy, and A.~L. Yuille, ``Deeplab:
  Semantic image segmentation with deep convolutional nets, atrous convolution,
  and fully connected crfs,'' \emph{IEEE Transactions on Pattern Analysis and
  Machine Intelligence}, vol.~40, no.~4, pp. 834--848, 2017.

\bibitem{gonzalez2018}
\BIBentryALTinterwordspacing
R.~Gonz{\'{a}}lez and K.~Iagnemma, ``Deepterramechanics: Terrain classification
  and slip estimation for ground robots via deep learning,'' \emph{CoRR}, vol.
  abs/1806.07379, 2018. [Online]. Available:
  \url{http://arxiv.org/abs/1806.07379}
\BIBentrySTDinterwordspacing

\bibitem{dimastrogiovanni2020}
\BIBentryALTinterwordspacing
M.~Dimastrogiovanni, F.~Cordes, and G.~Reina, ``Terrain estimation for
  planetary exploration robots,'' \emph{Applied Sciences}, vol.~10, no.~17,
  2020. [Online]. Available: \url{https://www.mdpi.com/2076-3417/10/17/6044}
\BIBentrySTDinterwordspacing

\bibitem{bai2019}
\BIBentryALTinterwordspacing
C.~Bai, J.~Guo, L.~Guo, and J.~Song, ``Deep multi-layer perception based
  terrain classification for planetary exploration rovers,'' \emph{Sensors},
  vol.~19, no.~14, 2019. [Online]. Available:
  \url{https://www.mdpi.com/1424-8220/19/14/3102}
\BIBentrySTDinterwordspacing

\bibitem{miliaresis2004}
\BIBentryALTinterwordspacing
G.~Miliaresis and N.~Kokkas, ``Segmentation and terrain modelling of
  extra‐terrestrial chasmata,'' \emph{Journal of Spatial Science}, vol.~49,
  no.~2, pp. 87--97, 2004. [Online]. Available:
  \url{https://doi.org/10.1080/14498596.2004.9635024}
\BIBentrySTDinterwordspacing

\bibitem{liang2017lunar}
J.~Liang and X.~Tian, ``A fast auto recognition algorithm for lunar terrain in
  wavelet domain,'' in \emph{2017 10th International Congress on Image and
  Signal Processing, BioMedical Engineering and Informatics (CISP-BMEI)}, 2017,
  pp. 1--6.

\bibitem{jiang2015}
\BIBentryALTinterwordspacing
H.-K. Jiang, X.-L. Tian, and A.-A. Xu, ``A new segmentation algorithm for lunar
  surface terrain based on {CCD} images,'' \emph{Research in Astronomy and
  Astrophysics}, vol.~15, no.~9, pp. 1604--1612, aug 2015. [Online]. Available:
  \url{https://doi.org/10.1088/1674-4527/15/9/016}
\BIBentrySTDinterwordspacing

\bibitem{zurn2021}
J.~Z{\"u}rn, W.~Burgard, and A.~Valada, ``Self-supervised visual terrain
  classification from unsupervised acoustic feature learning,'' \emph{IEEE
  Transactions on Robotics}, vol.~37, no.~2, pp. 466--481, 2021.

\bibitem{kahn2020}
\BIBentryALTinterwordspacing
G.~Kahn, P.~Abbeel, and S.~Levine, ``{BADGR:} an autonomous self-supervised
  learning-based navigation system,'' \emph{CoRR}, vol. abs/2002.05700, 2020.
  [Online]. Available: \url{https://arxiv.org/abs/2002.05700}
\BIBentrySTDinterwordspacing

\bibitem{holder2016}
C.~J. Holder, T.~P. Breckon, and X.~Wei, ``From on-road to off: Transfer
  learning within a deep convolutional neural network for segmentation and
  classification of off-road scenes,'' in \emph{Computer Vision -- ECCV 2016
  Workshops}, G.~Hua and H.~J{\'e}gou, Eds.\hskip 1em plus 0.5em minus
  0.4em\relax Cham: Springer International Publishing, 2016, pp. 149--162.

\bibitem{wagstaff2018}
\BIBentryALTinterwordspacing
K.~Wagstaff, Y.~Lu, A.~Stanboli, K.~Grimes, T.~Gowda, and J.~Padams, ``Deep
  mars: Cnn classification of mars imagery for the pds imaging atlas,''
  \emph{Proceedings of the AAAI Conference on Artificial Intelligence},
  vol.~32, no.~1, Apr. 2018. [Online]. Available:
  \url{https://ojs.aaai.org/index.php/AAAI/article/view/11404}
\BIBentrySTDinterwordspacing

\bibitem{sharma2019}
\BIBentryALTinterwordspacing
S.~Sharma, J.~E. Ball, B.~Tang, D.~W. Carruth, M.~Doude, and M.~A. Islam,
  ``Semantic segmentation with transfer learning for off-road autonomous
  driving,'' \emph{Sensors}, vol.~19, no.~11, 2019. [Online]. Available:
  \url{https://www.mdpi.com/1424-8220/19/11/2577}
\BIBentrySTDinterwordspacing

\bibitem{fernando2013}
B.~Fernando, A.~Habrard, M.~Sebban, and T.~Tuytelaars, ``Unsupervised visual
  domain adaptation using subspace alignment,'' in \emph{2013 IEEE
  International Conference on Computer Vision}, 2013, pp. 2960--2967.

\bibitem{gong2012}
B.~Gong, Y.~Shi, F.~Sha, and K.~Grauman, ``Geodesic flow kernel for
  unsupervised domain adaptation,'' in \emph{2012 IEEE Conference on Computer
  Vision and Pattern Recognition}, 2012, pp. 2066--2073.

\bibitem{ganin2015}
Y.~Ganin and V.~Lempitsky, ``Unsupervised domain adaptation by
  backpropagation,'' in \emph{Proceedings of the 32nd International Conference
  on International Conference on Machine Learning - Volume 37}, ser.
  ICML'15.\hskip 1em plus 0.5em minus 0.4em\relax JMLR.org, 2015, p.
  1180–1189.

\bibitem{coral2016}
B.~Sun and K.~Saenko, ``Deep coral: Correlation alignment for deep domain
  adaptation,'' in \emph{ECCV 2016 Workshops}, 2016.

\bibitem{ouali2020}
Y.~Ouali, C.~Hudelot, and M.~Tami, ``Semi-supervised semantic segmentation with
  cross-consistency training,'' in \emph{The IEEE/CVF Conference on Computer
  Vision and Pattern Recognition (CVPR)}, June 2020.

\bibitem{lv2020}
F.~Lv, T.~Liang, X.~Chen, and G.~Lin, ``Cross-domain semantic segmentation via
  domain-invariant interactive relation transfer,'' in \emph{Proceedings of the
  IEEE/CVF Conference on Computer Vision and Pattern Recognition (CVPR)}, June
  2020.

\bibitem{lambert2020}
J.~Lambert, Z.~Liu, O.~Sener, J.~Hays, and V.~Koltun, ``{MSeg}: A composite
  dataset for multi-domain semantic segmentation,'' in \emph{Computer Vision
  and Pattern Recognition (CVPR)}, 2020.

\bibitem{newell2017}
A.~Newell, Z.~Huang, and J.~Deng, ``Associative embedding: End-to-end learning
  for joint detection and grouping,'' in \emph{Proceedings of the 31st
  International Conference on Neural Information Processing Systems}, ser.
  NIPS'17.\hskip 1em plus 0.5em minus 0.4em\relax Red Hook, NY, USA: Curran
  Associates Inc., 2017, p. 2274–2284.

\bibitem{swan2021}
\BIBentryALTinterwordspacing
R.~M. Swan, D.~Atha, H.~A. Leopold, M.~Gildner, S.~Oij, C.~Chiu, and M.~Ono,
  ``{AI4MARS: A Dataset for Terrain-Aware Autonomous Driving on Mars},''
  \emph{2021 IEEE/CVF Conference on Computer Vision and Pattern Recognition
  Workshops (CVPRW)}, pp. 1982--1991, jun 2021. [Online]. Available:
  \url{https://ieeexplore.ieee.org/document/9523149/}
\BIBentrySTDinterwordspacing

\bibitem{he2015}
K.~He, X.~Zhang, S.~Ren, and J.~Sun, ``Deep residual learning for image
  recognition,'' \emph{arXiv preprint arXiv:1512.03385}, 2015.

\bibitem{sk2019}
X.~Li, W.~Wang, X.~Hu, and J.~Yang, ``Selective kernel networks,'' in
  \emph{Proceedings of the IEEE/CVF Conference on Computer Vision and Pattern
  Recognition}, 2019, pp. 510--519.

\bibitem{liang2017}
\BIBentryALTinterwordspacing
L.~Chen, G.~Papandreou, F.~Schroff, and H.~Adam, ``Rethinking atrous
  convolution for semantic image segmentation,'' \emph{CoRR}, vol.
  abs/1706.05587, 2017. [Online]. Available:
  \url{http://arxiv.org/abs/1706.05587}
\BIBentrySTDinterwordspacing

\end{thebibliography}

\end{document}